\documentclass{article}

\usepackage{arxiv}

\usepackage{doi}

\usepackage{algorithm}
\usepackage{algorithmic}

\usepackage{arxiv}

\usepackage[utf8]{inputenc} 
\usepackage[T1]{fontenc}    
\usepackage{hyperref}       
\usepackage{url}            
\usepackage{booktabs}       
\usepackage{amsfonts}       
\usepackage{nicefrac}       
\usepackage{microtype}      
\usepackage{lipsum}
\usepackage{fancyhdr}       
\usepackage{graphicx}       
\graphicspath{{figures/}}     

\title{Symbolic expression generation via Variational Auto-Encoder}

\author{ {\hspace{1mm}Sergei Popov} \\
	Department of Computer Science \\
        LAMBDA HSE University\\
        \\
        National University of Science and Technology MISIS\\
        \\
 Corresponding author : sapopov@edu.hse.ru\\
	\And
	{\hspace{1mm}} Mikhail Lazarev\\
	Department of Computer Science\\
	LAMBDA HSE University \\
	\And
	{\hspace{1mm}} Vladislav Belavin\\
	Department of Computer Science\\
	LAMBDA HSE University \\
 \And
	{\hspace{1mm}} Denis Derkach\\
	Department of Computer Science\\
	LAMBDA HSE University \\
 \And
	{\hspace{1mm}} Andrey Ustyuzhanin\\
	Department of Computer Science\\
	LAMBDA HSE University \\
 \\
        National University of Science and Technology MISIS\\
        \\
        Institute for Functional Intelligent Materials, National University of Singapore\\
        \\
        Constructor University, Switzerland\\
}

\makeatletter
\let\Title\@title
\makeatother

\begin{document}
\maketitle
\begin{abstract}

There are many problems in physics, biology, and other natural sciences in which symbolic regression can provide valuable insights and discover new laws of nature. A widespread Deep Neural Networks do not provide interpretable solutions. Meanwhile, symbolic expressions give us a clear relation between observations and the target variable. However, at the moment, there is no dominant solution for the symbolic regression task, and we aim to reduce this gap with our algorithm. In this work, we propose a novel deep learning framework for symbolic expression generation via variational autoencoder (VAE). In a nutshell, we suggest using a VAE to generate mathematical expressions, and our training strategy forces generated formulas to fit a given dataset. Our framework allows encoding apriori knowledge of the formulas into fast-check predicates that speed up the optimization process. We compare our method to modern symbolic regression benchmarks and show that our method outperforms the competitors under noisy conditions. The recovery rate of SEGVAE is 65\% on the Ngyuen dataset with a noise level of 10\%, which is better than the previously reported SOTA by 20\%. We demonstrate that this value depends on the dataset and can be even higher.
\end{abstract}

\newpage{}

\flushbottom{}
\maketitle{}
\thispagestyle{empty}

\section*{Introduction}
The discovery of new laws from experimental observations may seem to be an old and long-forgotten topic at first sight. Indeed, in the age of data-driven science, neural networks can easily fit a pretty complicated dependency. However, generalization and interpretation of those networks usually leaves much to be desired. For such phenomena as a biotech reaction, molecular dynamics potentials, or epidemic spread, a learning algorithm representing the dependency between target property and dependent characteristic in a simplistic and human-conceivable, such as usual formulas, is invaluable. There is also a lot of tasks where the black-box algorithms can't be trusted with your eyes closed, i.e. in self-driving cars, medicine, markets analyses, aircraft design and so on. In other words, areas where the cost of error is very high, so it's necessary to understand why algorithm made one decision or another. Such examples are everywhere. Moreover, the Symbolic Regressions techniques are suitable for solving an optimal control task \cite{SahooLampertMartius2018:EQLDiv}. Optimal control task deals with the problem of finding a control law for a given system such that a certain criterion is achieved, solution to this problem is a function of time \cite{DIVEEV2021646}.

Several methods enhancing the so-called symbolic regression approach have been developed recently \cite{AIFeynman}, where the goal was to reconstruct 100 formulas from Feynman's lectures on Physics. Also in resent paper symbolic regression was applied to find thermodynamic description of ionic transport out of experimental data \cite{D2DD00027J} the resulting formula reflects physical principles of underpinning ionic transport. Symbolic regression is akin to a simple regression, as it fits experimental data. However, symbolic regression tries to find suitable and functional feature transformations represented by a computational graph over the original feature vector. Such graphs impose additional complications while optimizing those using gradient descent-based approaches.

Another difficulty such methods regularly face is the inherent noisiness of experimental measurements. Hence, each algorithm should ideally provide theoretical or empirical guarantees of the noise level robustness. The example of noisy data can be found in all kind of experimental data and sometimes the noise level could be quite high depending on experiment type \cite{Eling2019} \cite{Reinbold2021}. Since the scientific theories or models has to be proven or validated by experiment the ability to reconstruct the symbolic solution of observe effect is critically important. We address this issue by the algorithm presented in this work.

\par One of our key contributions is the design of a novel process of the symbolic regression training SEGVAE that differs from the current state-of-the-art approaches: higher noise stability, higher data efficiency, and adjustability of the priors for the symbolic expression to the physics intuition of the user. Often scientists do know the frame of the searching formula (e.g., limits and approximations) and laws that data have to follow. We introduce a predicate mechanism for formula search and the ability to implement known conversation laws. These features are essential for processing experimental data.

The structure of this paper is the following: section 2 contains an overview of prior symbolic methods, and section 3 contains a description of our approach. All experiments with comparing performance are presented in section 4. Section 5 concludes the paper.

\section*{Literature Review}
The goal of using artificial intelligence to help discover the scientific laws underlying experimental data has been pursued in several works. Some of these works assume prior knowledge of the mystery environments of interest. However, the ones most relevant to our study are the ones that minimize any assumptions.

Since the symbolic regression problem is a discrete optimization task with the search space that exponentially depends on expression length, the majority of traditional approaches generally exploit genetic algorithms \cite{genetic-programming-sr} and \cite{koza-gp}. In this algorithms the process of optimization is inspired by the natural selection and relies on operators such as mutation, crossover and selection \cite{ea-michalewicz}. The new populations are produced by iterative applying of genetic aforementioned operators on individuals from current population. The most successful one of these approaches is the commercial software Eureqa (\cite{eureqa}), which was developed more than ten years ago and still holds one of the leading positions in the field.

There are several recent works dedicated to recovering physical laws in symbolic form. \cite{AIFeynman} introduce an AI Feynman algorithm and further improved in \cite{AIFeynman2} AI Feynman 2.0, which uses a) physics-inspired deep learning strategies, b) dimensional analysis like search for symmetries, separability, and alike, c) brute forces the simplified equation that recovers physical equations from experimental data. While this algorithm does a good job simplifying expressions, it struggles to recover expressions that could not be simplified enough.

PySR \cite{DBLP:journals/corr/abs-2006-11287} is basically reincarnation of Eureqa used friendly interface which allows to introduce predicates. PySR built on Julia, uses regularized evolution, simulated annealing, and gradient-free optimization and interfaced by Python. 

Interesting and intuitively easy approach was demonstrated in  \cite{DBLP:journals/corr/MartiusL16} later this method have been updated in \cite{pmlr-v80-sahoo18a} and in it's latest version \cite{https://doi.org/10.48550/arxiv.2105.06331}. The authors proposed architecture similar to a multilayer perceptron (MLP), where instead of a single activation for all outputs, they used a custom set of activation functions. The authors claim the efficiency of such an approach over MLP neural networks outside of the training set region. Thus good extrapolation capabilities have been demonstrated. But unfortunately, the authors did not present comparison results with other existing methods on common datasets.

Another deep learning approach to symbolic regression is introduced by \cite{dsr}. The authors present a gradient-based approach for symbolic regression based on reinforcement learning, which they call deep symbolic regression (DSR). DSR consists of a recurrent neural network that outputs a distribution over mathematical formulas. This network is used to sample equations, which will be evaluated based on the given dataset. Then the evaluation result will be used to further improve the distribution over mathematical formulas making the better expressions more probable. The DSR method has recently been updated by introducing a genetic programming component, significantly enhancing several benchmark tests. The latest algorithm version performs better than others to the best of our knowledge. Thus, we are using it (\cite{mundhenk2021seeding}) as a baseline known as DSO. DSO uses some prebuilt predicates to avoid nested repeating functions e.g. $sin(sin(sin(..)))$, but a more complex or physics-inspired predicates introduction is challenging. 

One of the recent manuscript was dedicated to mixed approach of computer vision methods and transformers to symbolic regression problem \cite{https://doi.org/10.48550/arxiv.2205.11798}. The workflow is the following: the input data is represented by image, this image is treated by convolution layers to create the image embedding to feed the transformer. However, proposed approach might be useful only for a datasets, containing a large number of points. In addition, the results of proposed algorithm was not compared with published SOTA aproaches.

SciNet (\cite{SciNet}) approach is inspired by human thinking along a physical modeling process. Just like human physicists do not rely on actual observations but rather on their compressed representation to make some theoretical conclusions, SciNet encodes the experimental data to a latent representation that stores different physical parameters and uses this representation to answer specific questions about the underlying physical system. Undoubtedly SciNet is successful at learning relevant physical concepts. However, its goals are very different from ours: it does not recover the laws in symbolic form but uses a neural network to model them.

In paper on Neural-Symbolic Regression that Scales \cite{pmlr-v139-biggio21a} authors propose to use pre-trained transformers \cite{NIPS2017_3f5ee243} to predict symbolic expression. The network consists of a transformer encoder and transformer decoder trained on generated formulas and BFGS algorithm for constants optimization. They compare transformer results to DSR (previous version of DSO) as a baseline. Despite good evaluation time, results quality is lower than DSR on the Nguyen dataset. 

Our method is akin to the work \cite{vae-rnn}. It adapts the variational autoencoder by using LSTM RNNs for both encoder and decoder. Thus, forming a sequence autoencoder with the Gaussian prior acting as a regularizer on the hidden code. The proposed generative model incorporates distributed latent representations of entire sentences. By examining paths through this latent space, it is possible to generate coherent novel sentences that interpolate between known sentences.


\section*{Methods}
This section introduces the Symbolic expression generation via Variational Auto-Encoder (SEGVAE) algorithm \cite{popov_sergei_2022_7364439}. In a nutshell, our architecture is a Variational Auto-encoder~\cite{kingma2013auto} in which the encoder and decoder are based on recurrent neural networks. The primary motivation behind using VAE for symbolic sequence representation is that VAE conveys a regularized learning method that minimizes the volume of low-energy (noise) representation space, thus preserving the richness of the signal (relevant sequences) representation. We also describe here our implementation of this architecture and the training procedures.

\subsection*{Architecture} \label{sect:method_architecture}
SEGVAE sequentially generates formulas. It is possible due to the one-to-one correspondence between sequences of tokens and formulas. Each token can be one of three types: input variables, constants, and operators. In this paper our typical library of operators is ['$add$', '$sub$', '$mul$', '$div$', '$sin$', '$cos$', '$log$', '$exp$']. The number of variables $X$ and constants depends on a task. We discuss the way to deal with constants in the subsection below. We use normal Polish notation to represent formulas as sequences in which operators precede their operands. The main advantage of Polish notation over the conventional one is that Polish representation is unambiguous and does not require brackets.

\textbf{Variational autoencoder.} VAE is a generative encoder-decoder based latent variable model. Given an observation space $\mathcal{X}$ with a distribution $p(x)$ the model’s encoder maps it into a latent space $\mathcal{Z}$ with a distribution $p(z)$, and the model’s decoder maps $\mathcal{Z}$ back into the observation space $\mathcal{X}$.
Let $m$ - dimensionality of the latent space, $\mu_i$, $\sigma_i^2$ - the $i^{th}$ components of vectors $\mu(x), \sigma^2(x)$. If $p(z) = \mathcal{N}(0,1)$ and $q_{\theta_E}(z|x) = \mathcal{N}(\mu(x), \sigma^2(x))$, then the objective takes the following form:

\begin{equation}\label{first_eq}
-KL(q_{\theta_E}(z|x)||p(z)) + \mathbb{E}_{q_{\theta_E}(z|x)}\log p_{\theta_D}(x|z)
\end{equation}

This approach allows the model to decode plausible equations from every point in the latent space that has a reasonable probability under the prior. As long as we can represent expressions as sequences of tokens, we rely on traditional NLP approaches. Our encoder and decoder are both one-layer LSTMs with 64 hidden units.

\begin{figure*}[h]
	\noindent
	\centering
	\includegraphics[width=10.6cm]{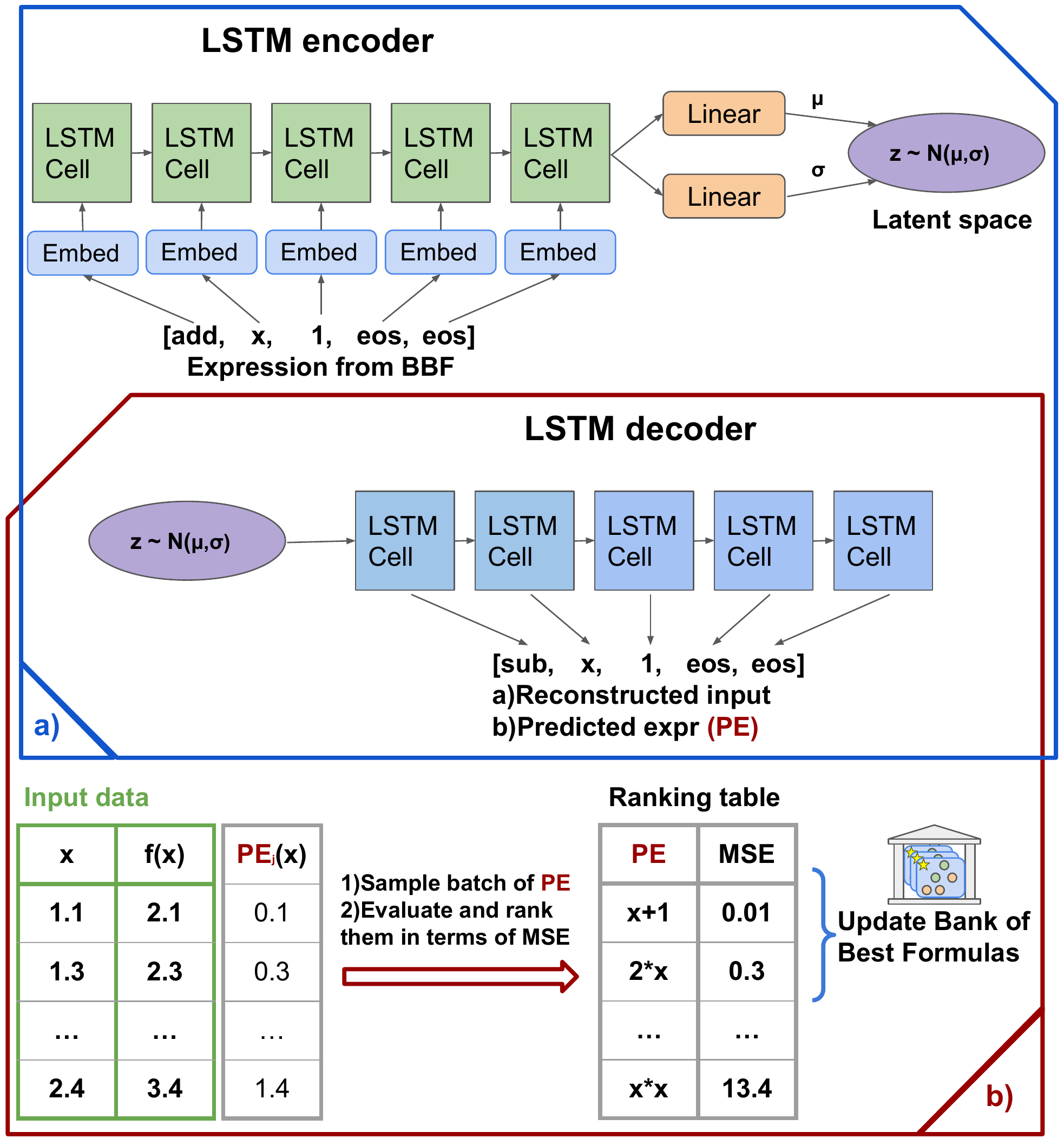}
	\caption{The SEGVAE architecture and training scheme.(a) Pretraining and training scheme. (b) Sampling scheme, where output formula evaluates and goes to the Bank of Formulas }
	\label{fig:architecture}
\end{figure*}

\begin{figure*}[h]
\centering\includegraphics[width=14.6cm]{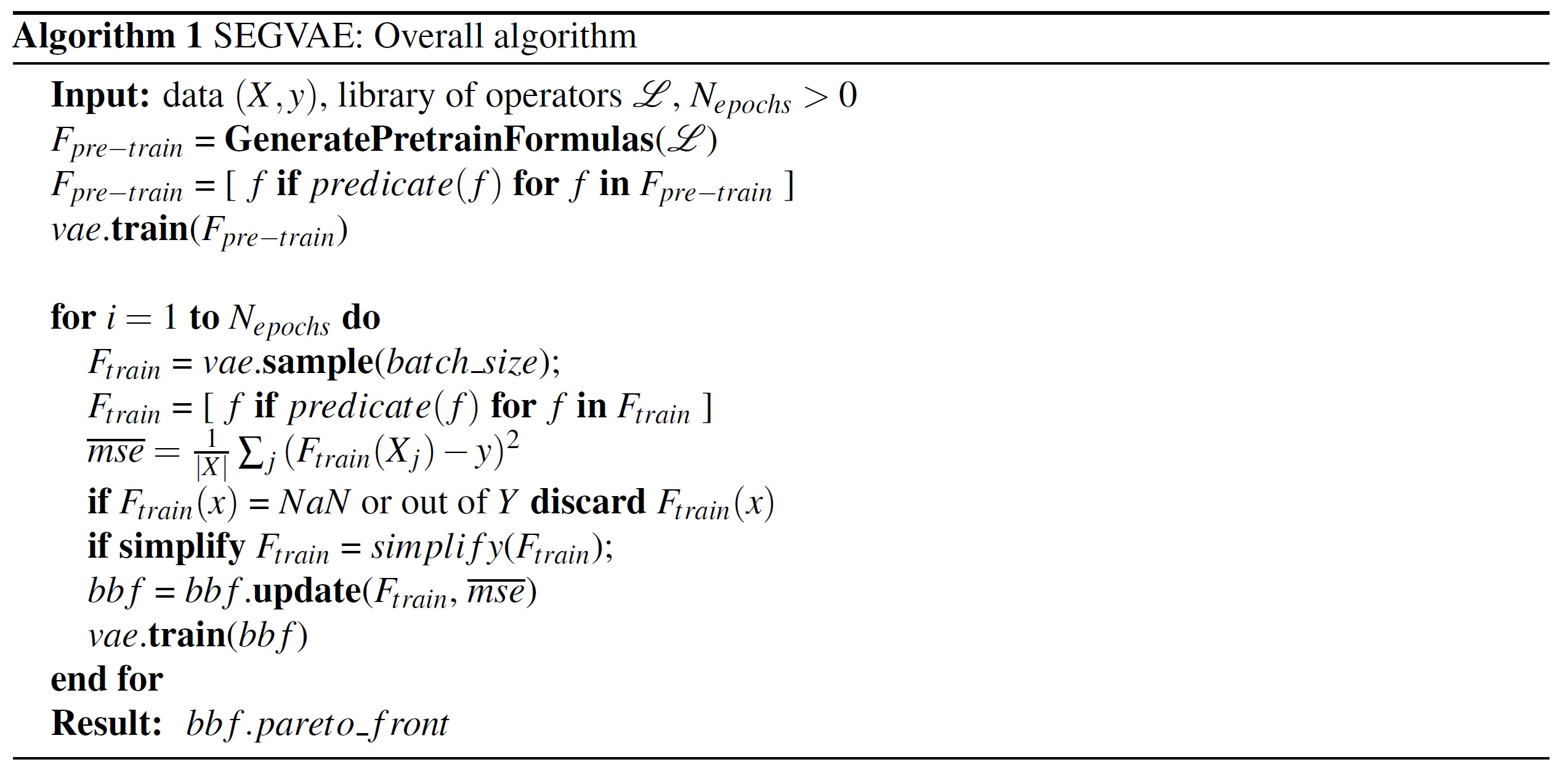}
\end{figure*}

\subsection*{Training procedure}\label{sect:method_training}
Here we summarize the details of our training protocol, which consists of two steps pre-training and the main training cycle.

\textbf{Pre-training.} This step allows the model to memorize the general formula structure and generate valid formulas afterward. Firstly, we randomly sample sequences of tokens from given library $\mathcal{L}$. However, uniformly sampling expressions with $n$ internal nodes is not a simple task. Naive algorithms tend to favor specific kinds of expressions. So, we follow the data generator technique introduced in \cite{DL4SM}. Secondly, we choose only statements that meet our predicate conditions and thus create a pre-train dataset. Also, we add to the pre-train dataset a set of generalized formulas that commonly appear in natural science. Then the variational autoencoder is trained on these formulas. As a result of training, the vast majority of generated formulas in the next steps of the algorithm are valid formulas, so for the sake of simplicity, we can safely ignore invalid formulas in the following stages. Note that this step does not depend on the target task at all, so we do the pre-training once for each library $\mathcal{L}$. The schematic picture of pre-train process and algorythm arhitecture presented in Figure 1 (a). The Figure 1 (b) shows the formation of the Bank of Best Formulas.

\textbf{Main training cycle.} When the model can generate valid formulas, it is time to teach it to generate valid formulas which describe a given environment or system under exploration. Firstly, a batch of formulas is sampled using the variational autoencoder for each epoch. Then all the duplicates and invalid formulas are removed. Secondly, each formula $f$ is evaluated on the dataset $\mathcal{D} = (X_d, y_d)$ in terms of mean squared error between $f(X_d)$ and $Y_d$:

\begin{equation}\label{second_eq}
error(f) = \frac{1}{|\mathcal{D}|} \sum_{x, y \in \mathcal{D}}{(f(x) - y)^2}
\end{equation}

Then the $P$ percent of the formulas with the smallest mean squared error are candidates to be saved to the bank of the best formulas (BBF). First, those candidates needs to be checked on correct  of definition and values, by default $x \in (X_{min},X_{max})$ and $y \in (Y_{min},Y_{max})$ defined by the dataset. We evaluate function $f(x)$ on points sampled from a uniform distribution on $(X_{min},X_{max})$ and compute $\hat{y}=f(x)$ for those points. If $\hat{y}$ gets out of the $(Y_{min},Y_{max})$ region or we get a $NaN$, we discard this formula. This approach does not guarantee that the function $f$ is defined on a given domain, e.g., it cannot find discontinuities. However, as we show in our experiments, this approach improves the performance of our model, and it is computationally efficient compared to analytical evaluation. Second, if needed, formulas can be simplified using the $sympy$ library before saving it in BBF. This bank keeps formulas from the last $N$ epochs. Our typical value for both hyperparameters $P$ and $N$ is $20$ and $5$. Finally, the VAE is fine-tuned on formulas from the BBF. Details of the network training are specified in the supplementary materials. Training overview presented in Figure 2.

\begin{figure}[h]
	\noindent
	\centering
	\includegraphics[width=10.6cm]{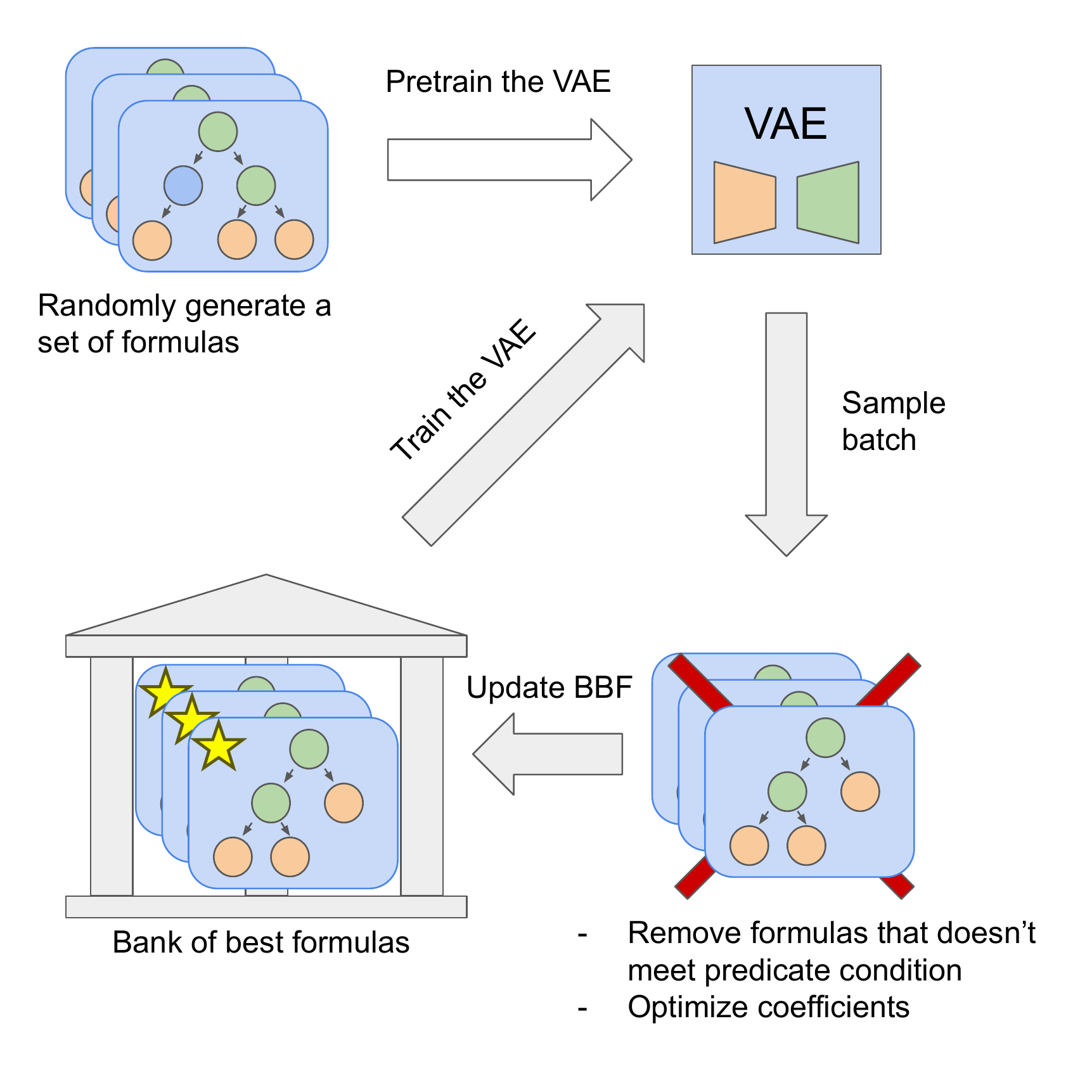}
	\caption{The SEGVAE algorithm training scheme. Pre-training stage and main training cycle.}
	\label{fig:algo}
\end{figure}

\textbf{Constants.} There are two ways of dealing with constants in the resulting formulae. The first method supposes that all constants are incorporated in a library $\mathcal{L}$. In this case, constants are regular tokens, and we do not need any modifications to our algorithm. The main drawback of this approach is the lack of expressiveness. However, it significantly helps the algorithm avoid overfitting to noisy data.

The second method is generating placeholders for future constants by including token $'const'$ to the library $\mathcal{L}$. Then, after the sampling stage, each placeholder is replaced by a parameter which we minimize the mean squared error between $f(X_d, consts)$ and $Y_d$ by BFGS optimization algorithm (\cite{bfgs}).

\textbf{Predicates.} It is relatively straightforward to incorporate some prior knowledge about the target formula in the proposed algorithm framework. At the training stage, while we choose the best-sampled formulas by VAE, we ignore formulas that do not meet initial conditions regardless of their metrics. This technique reduces the search space, which helps SEGVAE find correct equations, as demonstrated in the experiment section below.

Imagine we know that the ratio of polynomials describes our physical system dynamics well, and the task is to determine that exact dependency. One way of helping the model is to reduce search space by shrinking the library of operators, e.g., excluding trigonometrical operations. However, one can go further, creating a condition on possible positions for remaining tokens. In our case ['$div$'] operator should be located only in the first place. Eventually model will learn not to generate formulas that violate prior conditions. We will demonstrate algorithm work results on chosen list of formulas and predicates in this paper below. The list of formulas and related predicates are listed in Table 3.


\subsection*{Inference}\label{sect:method_inference} 
Once the VAE model is trained, one can sample a batch of candidate expressions that are supposed to fit the given dataset. Overwhelmed formulas that describe the dataset best in terms of mean squared error could be overfitted to the noise in the data. So, there is a trade-off between error and complexity. To choose final expressions, we evaluate the complexity of each equation as:

\begin{equation}\label{third_eq}
C(t)= \sum_{i}^{T} c(\gamma_i) 
\end{equation}

where $c$ is complexity of given token, which is equal to one for all input variables, constants and operators ['$add$', '$mul$', '$sub$'], $c$ is two for ['$div$'], three for ['$sin$', '$cos$'] and finally four for ['$log$', '$exp$']. Then we use this $C$ and error values to identify Pareto-frontier and allow user to pick those formulas that satisfy her needs.

\section*{Results}
This section reports comparison results between our approach and state-of-the-art symbolic regression packages. We took the bests of our knowledge algorithms, namely Deep Symbolic Optimization (DSO) \cite{mundhenk2021seeding}, since the DSO superior performance was supplemented by Figure 3. We have used Nguyen 1-12 formulas (Table 1) and formulas listed in table 3 to generate datasets. We compare SEGVAE results with those of the DSO algorithm we take from the github repository.   

 \begin{figure}[h]
	\noindent
	\centering
	\includegraphics[width=8.8cm]{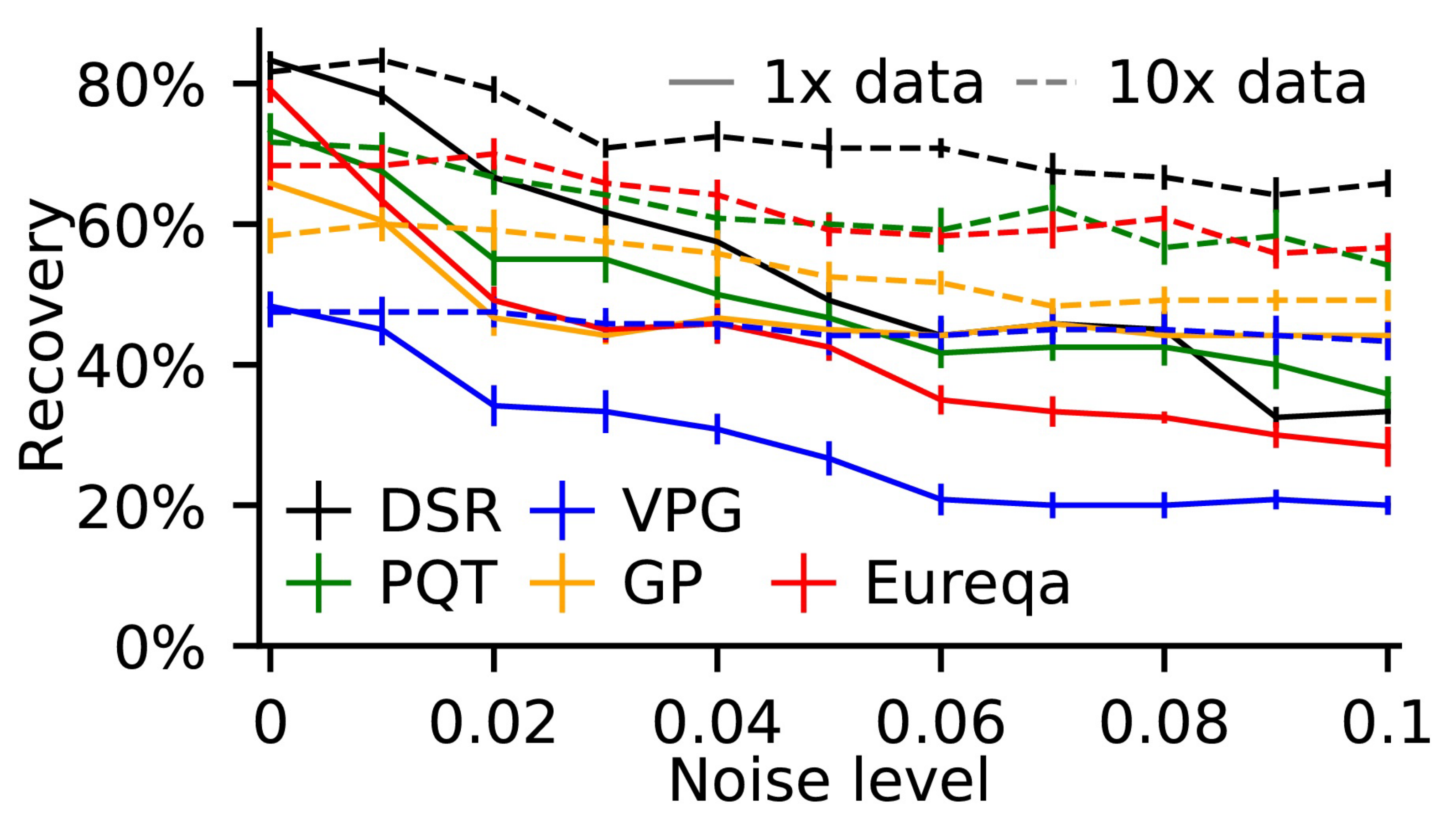}
	\caption{Recovery rate vs dataset noise and dataset size across all Nguyen benchmarks.Error bars represent standard error. (adopted from \cite{mundhenk2021seeding})}
	\label{fig:algo}
\end{figure}

\textbf{Datasets.}   Each symbolic regression task corresponds to a table of numbers, those rows are of the form $x_1, .., x_n, y$, where $y = f(x_1, .., x_n)$. The task is to discover the correct symbolic expression $f$.

\begin{table*}[h]
    \caption{Nguyen Dataset. Variables are denoted as $x_1$ and $x_2$. Variables are uniformly sampled, U(a, b, c) denotes $c$ times sampling between $a$ and $b$ for each input variable, $N$ natural numbers, $L_0 = [add, sub, mul, div, exp, ln, sin, cos]$.}
    \label{sample-table}
    \centering
    \begin{tabular}{lllll}
        \toprule
        Name     & Expression & Dataset  & Library \\
        \midrule
        Nguyen-1 & $ x_1^3 + x_1^2 + x_1 $ & $U(-1, 1, 20)$   & $L_0 + [0.5, 1, -1, 2]$ \\
        Nguyen-2 & $ x_1^4 +x_1^3 + x_1^2 + x_1 $ & $U(-1, 1, 20)$  & $L_0 + [0.5, 1, -1, 2]$  \\
        Nguyen-3 & $ x_1^5 + x_1^4 +x_1^3 + x_1^2 + x_1 $ & $U(-1, 1, 20)$  & $L_0 + [0.5, 1, -1, 2]$ \\
        Nguyen-4 & $ x_1^6 + x_1^5 + x_1^4 +x_1^3 + x_1^2 + x_1 $ & $U(-1, 1, 20)$  & $L_0 + [0.5, 1, -1, 2]$  \\
        Nguyen-5 & $ sin(x_1^2 )cos(x_1)-1 $ & $U(-1, 1, 20)$  & $L_0 + [0.5, 1, -1, 2]$  \\
        Nguyen-6 & $ sin(x_1) + sin(x_1 + x_1^2 ) $ & $U(-1, 1, 20)$  & $L_0 + [0.5, 1, -1, 2]$  \\
        Nguyen-7 & $ log(x_1 + 1) + log(x_1^2 + 1) $ & $U(0, 2, 20)$  & $L_0 + [0.5, 1, -1, 2]$ \\
        Nguyen-8 & $ sqrt(x_1) $ & $U(0, 4, 20)$  & $L_0 + [0.5, 1, -1, 2]$  \\
        Nguyen-9 & $ sin(x_1) + sin(x_2^2) $ & $U(0, 1, 20)$  & $L_0 + [0.5, 1, -1, 2]$  \\
        Nguyen-10 & $ 2sin(x_1)cos(x_2) $ & $U(0, 1, 20)$  & $L_0 + [0.5, 1, -1, 2]$  \\
        Nguyen-11 & $ x_1^{x_2} $ & $U(0, 1, 20)$  & $L_0 + [0.5, 1, -1, 2]$  \\
        Nguyen-12 & $ x_1^4 - x_1^3 + 0.5x_2^2 - x_2 $ & $U(0, 1, 20)$  & $L_0 + [x^N,0.5,N, e, pi ]$  \\
        \bottomrule
    \end{tabular}

\end{table*}

\begin{itemize}
	\item \textbf{Nguyen.} Nguyen benchmark is commonly used as a symbolic regression benchmark. Nguyen dataset consists of 12 formulas. We used the same dataset as in the \cite{dsr} and its updates (see table 1). To check the method's robustness, we add Gaussian noise proportional to $y$.

	\item \textbf{Livermore.} New benchmark dataset from the authors of \cite{mundhenk2021seeding} and \cite{dsr}. It consists of 22 formulas. However, we do not take all of them due to their apparent similarity to the Nguyen dataset.
\end{itemize}

\textbf{Ablation studies.} As was described above, SEGVAE has many parameters such as the number of layers, sampled formulas in pretrained step, number of an epoch, and a maximum length of generated expression. To find optimal parameters, we have performed an ablation study. As a baseline to tune SEGVAE parameters, we have used a subset of the Ngyuen dataset, namely Ngyuen-4,5,9,10. First, we concluded that a maximum expression length of 30 is perfectly balanced in terms of model expressibility and model stability. Secondly, we found that hidden dimensionality in the range of 64 to 256 does not affect the recovery rate of 50\%. By \textit{recovery} here, we mean exact symbolic equivalence of the suggested expression to the original formula. For example, we initialize the algorithm 100 times with different random seeds for a given dataset generated by some formula. Out of 100 runs, 80 correctly represented the initial formula. In this case recovery rate is $80\%$.
Nevertheless, the recovery rate depends on latent space, as presented in table 2. We have found that latent configuration space of size 128 with 128 hidden units to be optimal for this study. Another critical observation is the SEGVAE's recovery rate dependence on the Library selection. We checked the dependence of the recovery rate on the number of tokens in the library for the DSO and SEGVAE algorithms. Small library size may be why the algorithm cannot find a correct formula. It is simply because not enough tokens are available to describe a formula.
On the other hand, an over-inflated library exponentially increases algorithm search space for limited search iteration numbers. Thus, choosing excessive library contents may be a reason for a miserable formula reconstruction. A scientist has some prior knowledge about unknown yet dependency in an actual research process. Thus, we can use both predicates and a task-related library to simulate an actual searching formula situation. In the last series of ablation experiments, we show that our proposed method of discarding formulas that do not satisfy a given domain improves algorithm convergence speed by 50\% for the Nguyen dataset. Moreover, our experiments show that it enhances the recovery rate of Livermore-5 and Livermore-7 equations by 100\%

\textbf{Noise in data.} 
Noisy data were created by adding Gaussian noise with zero mean and standard deviation proportional to the root-mean-square of the dependent variable $y$. To check the models' robustness, we present averaged recovery rates as a function of the noise from $0\%$ (noiseless) to the maximum $10\%$. The main difficulty of regression with noise is the model's tendency to overfit the data. Sometimes increasing the number of points in a dataset may help.

\begin{table}
    \caption{Mean recovery rate dependence from latent space dimension with fixed hidden dimensions on sub Nguyen Dataset.}

    \label{sample-table}
    \centering
    \begin{tabular}{lllllll}
        \toprule

        Latent space dimension & $8 $ & $16$ & $32$  &  $64$ & $128$ & $256$ \\
        Mean recovery rate  & $50 $ & $58$ & $60$  &  $65$ & $75$ & $75$  \\
        
        \bottomrule
        
    \end{tabular}
\end{table}

\textbf{Experiments.}
We have evaluated the SEGVAE algorithm on the most commonly used Nguyen benchmark, consisting of 12 formulas. We compare and evaluate our models on two variants of Nguyen datasets. For the first ($Dataset$ $I$) variant, we sample only 20 uniformly distributed points, as shown in table 1. We have used the same hyperparameters in SEGVAE for all runs (see details in the supplementary materials). The number of examined expressions is set to 2 million per run, the same as in the DSO paper (\cite{mundhenk2021seeding}). 

\begin{figure}[h]
	\noindent
	\centering
	\includegraphics[width=8.5cm]{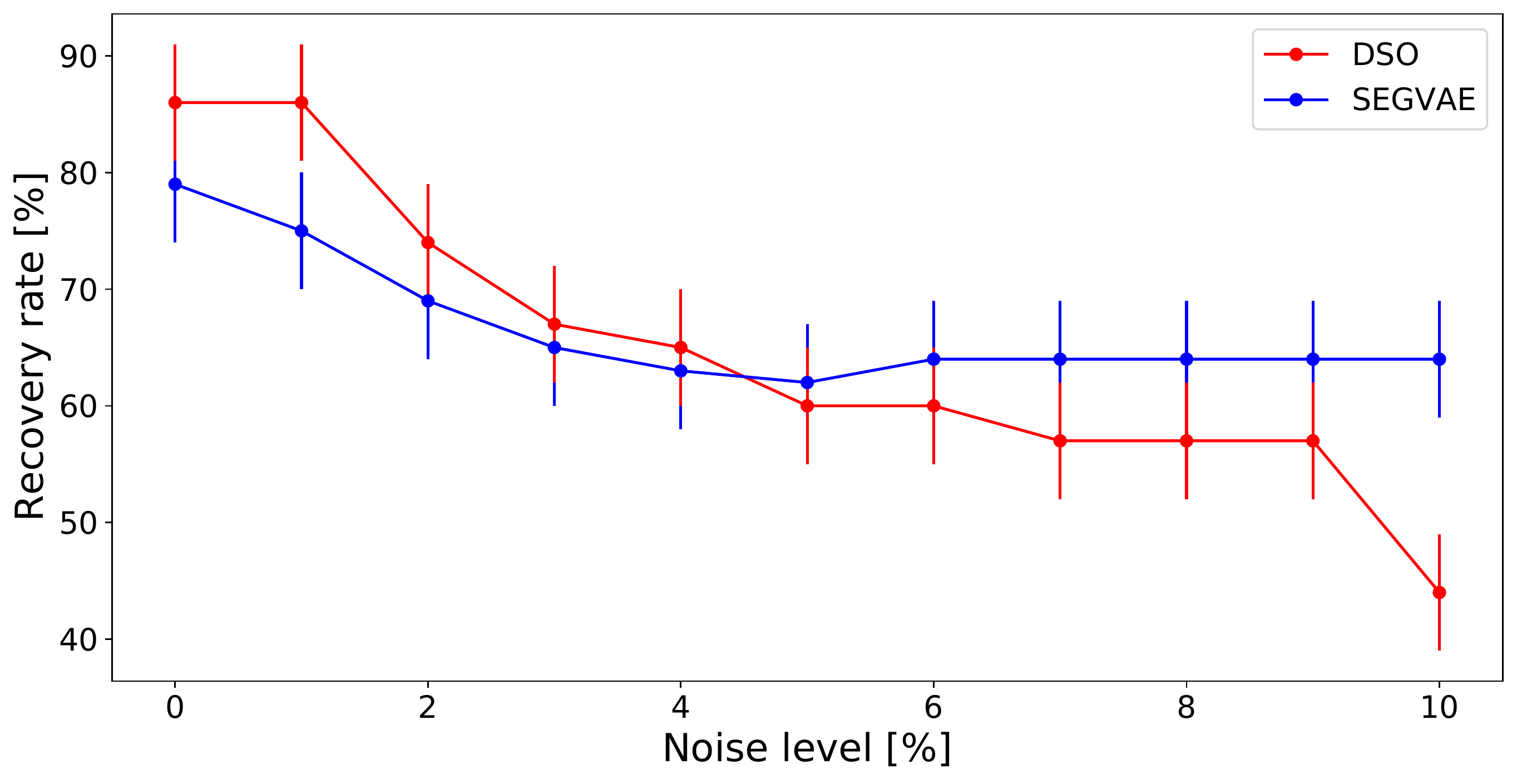}
	\caption{Average recovery rates of SEGVAE (blue) and DSO (red) algorithms on Ngyuen dataset with error bars.}
	\label{fig:algo}
\end{figure}
 
A comparison of DSO and SEGVAE on Dataset is presented in Figure 4 as the dependence of averaged recovery rates over noise level. Red and blue colors denote outputs of DSO and SEGVAE algorithms correspondingly. Essential to reiterate that by recovery, we mean at least one exact match with the wanted formula in the Pareto frontier.

\begin{table*}[h]
    \caption{ List of formulas and used predicates. Variables are denoted as $x$ and $y$. Variables are uniformly sampled, U(a, b, c) denotes $c$ times sampling between a and b for each input variable, $N$ natural numbers, $L_0 = [add, sub, mul, div, exp, ln, sin, cos]$.}
    \label{sample-table}
    \centering
    \renewcommand{\arraystretch}{1.5}
    \begin{tabular}{lllll}
        \toprule
        Formula name & Formula & Predicate & Dataset & Library  \\
        \midrule
        Nguyen-12 & $ x_1^4 - x_1^3 + \frac{1}{2}x_2^2 - x_2 $ & $ f(x_1) + g(x_2) $ & $U(0, 10, 200)$  & $L_0+[x^N,0.5, N, e, pi ]$  \\
        Neat-8 & $ \frac{exp(-(x_1-1)^2)}{1.2+(x_2-2.5)^2} $ & $ exp(f(x_1))/(g(x_2)) $ & $U(0.3, 4, 100)$  & $L_0+[0.5, 1, -1, 2]$  \\
        Neat-9 & $ \frac{1}{1+x_1^4} + \frac{1}{1+x_2^4}$ & $ 1/f(x_1) + 1/g(x_2) $ & $U(-5, 5, 21)$  & $L_0+[0.5, 1, -1, 2]$  \\
        Livermore-5 & $ x_1^4 - x_1^3 + x_1^2 - x_2 $ & $ f(x_1) + g(x_2) $ & $U(0, 1, 20)$  & $L_0+[0.5, 1, -1, 2]$  \\
        Livermore-7 & $ \frac{1}{2}e^{x_1} - \frac{1}{2}e^{-x_1} $ & $ f(x_1) - 1/g(x_1) $ & $U(-1, 1, 20)$ & $L_0+[0.5, 1, -1, 2]$   \\
        Livermore-8 & $ \frac{1}{2}e^{x_1} + \frac{1}{2}e^{-x_1} $ & $ f(x_1) + 1/g(x_1) $ & $U(-1, 1, 20)$  & $L_0+[0.5, 1, -1, 2]$  \\
        Livermore-10 & $ 6sin(x_1)cos(x_2) $ & $ const*f(x_1)*g(x_2) $ & $U(0, 1, 20)$  & $L_0+[x^N,0.5, N, e, pi ]$  \\
        Livermore-17 & $ 4sin(x_1)cos(x_2) $ & $ const*f(x_1)*g(x_2)  $ & $U(0, 1, 20)$ & $L_0+[x^N,0.5, N, e, pi ]$   \\
        Livermore-22 & $ exp(-\frac{1}{2}x_1^2) $ & $ exp(f(x_1)) $ & $U(0, 1, 20)$  & $L_0+[0.5, 1, -1, 2]$  \\
        R-2* & $ \frac{(x_1 + 1)^3}{x_1^2 - x_1 + 1} $ & $ f(x_1)/g(x_1) $ & $U(-10, 10, 20)$  & $L_0+[0.5, 1, -1, 2]$  \\
        
        \bottomrule
    \end{tabular}
\end{table*}

SEGVAE and DSO show similar average recovery rates (ARR) with below moderate noise levels. We average the recovery rate overall Nguyen formulas at a given noise level to compute ARR. With increasing noise levels, DSO recovery rates slowly go down. On the other hand, SEGVAE demonstrates good recovery rates stability up to the noise level of 10\% with a recovery rate of 70\%. The difference becomes visible at high noise levels where SEGVAE slightly outperforms DSO, 70\% vs. 45\% at maximum noise level. SEGVAE algorithm demonstrates higher noise stability on this dataset even without prior knowledge.

\begin{table*}[!b]
    \caption{Mean recovery rate dependence from latent space dimension with fixed hidden dimensions on sub Nguyen Dataset.}
    \label{sample-table}
    \centering
    \begin{tabular}{lllllllllll}
        \toprule
         & $N-12$ & $Neat-8$ & $Neat-9$ & $L-5$ & $L-7$\\
        \midrule
        SEGVAE & $100\%$  & $0\%$   & $0\%$   & $60\%$   & $20\%$  \\
        DSO & $0\%$  & $0\%$   & $0\%$   & $80\%$   & $0\%$ \\
        \midrule
        & $L-8$ & $L-10$ & $L-17$ & $L-22$ & $R-2*$ \\
        \midrule
        SEGVAE & $0\%$    & $100\%$   & $100\%$   &   $100\%$   & $0\%$ \\
        DSO & $0\%$    & $63\%$   & $57\%$   & $84\%$   & $4\%$ \\

        \bottomrule
    \end{tabular}
\end{table*}

 We compared algorithms and selected formulas from the original DSO paper, where the DSO algorithm shows daunting results. These formulas are presented in Table 3. This time we used specific predicates and token libraries to reveal these formulas and, at the same time to demonstrate the power of our approach. The results of this comparison are summarised in the same way through the ARR in Figure 5. The detailed results on noiseless data are presented in Table 4.

It often appears that scientists already know the general form of the object of interest. Thus, it would be natural to add predicates to let SEGVAE search formulas in a specific domain. We have seen that our VAE-based algorithm gives a similar result to the recent DSO results on the Nguyen dataset without any prior knowledge. However, there are many equations in which DSO is not accurate enough. To demonstrate the power of predicates in SEGVAE, we took predicates as listed in Table 3 and compared our results to DSO. The comparison results on noiseless data are presented in Table 4. The SEGVAE approach with predicates demonstrates a superior recovery rate than DSO, especially on noiseless data. More detailed results on noiseless are presented in Table 4. The proper predicates and optimal library play a crucial role in this case. In reality, scientists do not know the exact functional form of the studied effect. We can not use the recovery rate as a benchmark in this case. The only benchmark we can trust is MSE and formula shape or predicates that carry common scientific sense.
 
 \begin{figure*}[h]
	\noindent
	\centering
	\includegraphics[width=8.5cm]{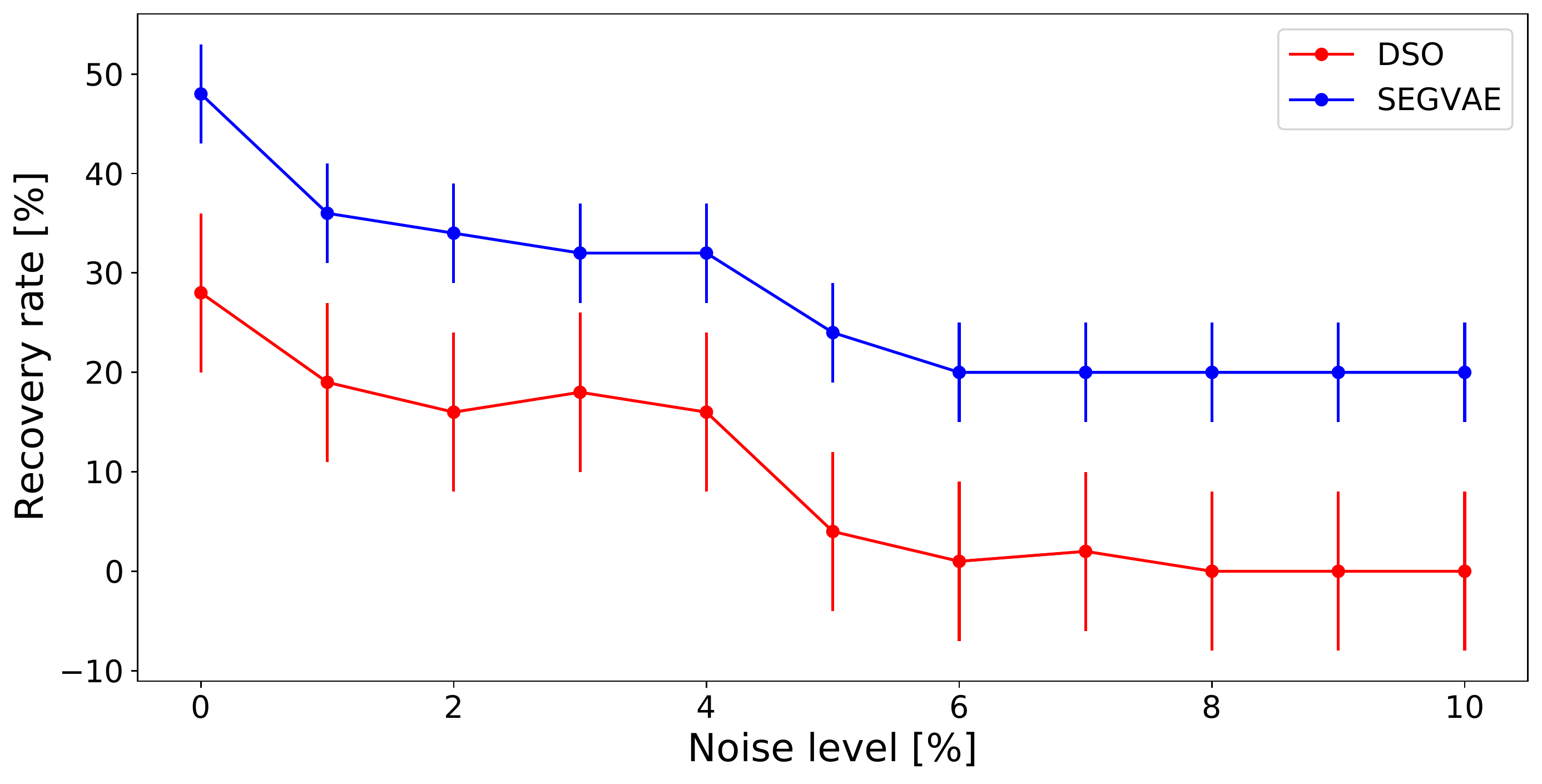}
	\caption{Average recovery rates of SEGVAE (blue) and DSO (red) algorithms based on formulas listed in table 3 with error bars.}
	\label{fig:algo}
\end{figure*}



\section*{Conclusions}
We introduce a novel algorithm, SEGVAE, for searching for symbolic representation of functional dependence from dependent variables. This approach is based on the VAE generative model that produces mathematical expressions and is constrained by apriori knowledge encoded in the form of fast-check predicates. Those predicates can express, for example, allowed formula patterns, domain, and possible output intervals.

As a benchmark, we have used a set of formulas introduced in the DSO paper, namely the Nguyen and the Livermore datasets. Our approach has the flexibility of formulating a priory physical knowledge in the form of a) library of functions, b) pre-training functions, and c) selection predicates used for pruning incorrectly generated expressions. Besides the sole accuracy of the method, we have focused on the algorithm's performance in realistic noisy environments. Symbolic regression approaches excel significantly compared to deep learning models where interpretability is paramount. Thus, the demand for such approaches is high in such scientific branches as material sciences, biotechnologies, and astrophysics, to mention a few.

We have systematically compared SEGVAE with DSO and shown superior performance of our approach, thus outperforming Eureqa, Wolfram, and alike, which DSO has dominated before. For the scarce-data regime and high-noise regimes, the SEGVAE significantly outperforms the competitors. However, this approach has its' limitations: a) the output formula's quality depends on the dataset and its size, b) obtaining adequate formula may require the user's assistance, namely in choosing predicates and the final formula from Pareto-front, and c) running time may be noticeable, i.e., around tens of minutes on a modern GPU. 

We pointed out the importance of the library size and showed that SEGVAE discovered formulas unreachable for DSO thanks to the flexibility of predicates supported by our method. The recovery rate
of SEGVAE improves the previously reported SOTA by 20\%. Since experimental data usually contains noise and some prior knowledge on the functional dependency is typically available, the SEGVAE benefits can be easily seen from an application point of view. Therefore, our model can come in handy in practical cases where interpretable symbolic solutions are needed to understand processes underlying experimental observations.
The code is available at the link: $https://github.com/anonnipsuser/segvae$

\bibliographystyle{unsrt}
\bibliography{sample}  

\begin{thebibliography}{10}

\bibitem{SahooLampertMartius2018:EQLDiv}
Subham~S. Sahoo, Christoph~H. Lampert, and Georg Martius.
\newblock Learning equations for extrapolation and control.
\newblock In {\em Proc. \textbackslash 35th International Conference on Machine
  Learning, {ICML} 2018, Stockholm, Sweden, 2018}, volume~80, pages 4442--4450.
  {PMLR}, 2018.

\bibitem{DIVEEV2021646}
A.I. Diveev, S.V. Konstantinov, and A.M. Danilova.
\newblock Solution of the optimal control problem by symbolic regression
  method.
\newblock {\em Procedia Computer Science}, 186:646--653, 2021.
\newblock 14th International Symposium "Intelligent Systems.

\bibitem{AIFeynman}
Silviu-Marian Udrescu and Max Tegmark.
\newblock Ai feynman: A physics-inspired method for symbolic regression.
\newblock {\em Science Advances}, 6(16), 2020.

\bibitem{D2DD00027J}
Eibar Flores, Christian Wölke, Peng Yan, Martin Winter, Tejs Vegge, Isidora
  Cekic-Laskovic, and Arghya Bhowmik.
\newblock Learning the laws of lithium-ion transport in electrolytes using
  symbolic regression.
\newblock {\em Digital Discovery}, 1:440--447, 2022.

\bibitem{Eling2019}
Nils Eling, Michael~D. Morgan, and John~C. Marioni.
\newblock Challenges in measuring and understanding biological noise.
\newblock {\em Nature Reviews Genetics}, 20(9):536--548, Sep 2019.

\bibitem{Reinbold2021}
Patrick A.~K. Reinbold, Logan~M. Kageorge, Michael~F. Schatz, and Roman~O.
  Grigoriev.
\newblock Robust learning from noisy, incomplete, high-dimensional experimental
  data via physically constrained symbolic regression.
\newblock {\em Nature Communications}, 12(1):3219, May 2021.

\bibitem{genetic-programming-sr}
Dominic~P. Searson, David~E. Leahy, and Mark~J. Willis.
\newblock Gptips: An open source genetic programming toolbox for multigene
  symbolic regression.
\newblock 2010.

\bibitem{koza-gp}
John~R. Koza.
\newblock Genetic programming as a means for programming computers by natural
  selection.
\newblock {\em Statistics and computing}, 4(2):87--112, 1994.

\bibitem{ea-michalewicz}
Zbigniew Michalewicz and Marc Schoenauer.
\newblock {Evolutionary Algorithms for Constrained Parameter Optimization
  Problems}.
\newblock {\em Evolutionary Computation}, 4(1):1--32, 03 1996.

\bibitem{eureqa}
Michael Schmidt and Hod Lipson.
\newblock Distilling free-form natural laws from experimental data.
\newblock {\em Science}, 324(5923):81--85, 2009.

\bibitem{AIFeynman2}
Silviu-Marian Udrescu, Andrew Tan, Jianhai Feng, Orisvaldo Neto, Tailin Wu, and
  Max Tegmark.
\newblock Ai feynman 2.0: Pareto-optimal symbolic regression exploiting graph
  modularity.
\newblock In {\em Advances in Neural Information Processing Systems 33
  pre-proceedings (NeurIPS 2020)}, 12/2020 2020.

\bibitem{DBLP:journals/corr/abs-2006-11287}
Miles~D. Cranmer, Alvaro Sanchez{-}Gonzalez, Peter~W. Battaglia, Rui Xu, Kyle
  Cranmer, David~N. Spergel, and Shirley Ho.
\newblock Discovering symbolic models from deep learning with inductive biases.
\newblock {\em CoRR}, abs/2006.11287, 2020.

\bibitem{DBLP:journals/corr/MartiusL16}
Georg Martius and Christoph~H. Lampert.
\newblock Extrapolation and learning equations.
\newblock {\em CoRR}, abs/1610.02995, 2016.

\bibitem{pmlr-v80-sahoo18a}
Subham Sahoo, Christoph Lampert, and Georg Martius.
\newblock Learning equations for extrapolation and control.
\newblock In Jennifer Dy and Andreas Krause, editors, {\em Proceedings of the
  35th International Conference on Machine Learning}, volume~80 of {\em
  Proceedings of Machine Learning Research}, pages 4442--4450. PMLR, 10--15 Jul
  2018.

\bibitem{https://doi.org/10.48550/arxiv.2105.06331}
Matthias Werner, Andrej Junginger, Philipp Hennig, and Georg Martius.
\newblock Informed equation learning, 2021.

\bibitem{dsr}
Brenden~K Petersen, Mikel~Landajuela Larma, Terrell~N. Mundhenk, Claudio~Prata
  Santiago, Soo~Kyung Kim, and Joanne~Taery Kim.
\newblock Deep symbolic regression: Recovering mathematical expressions from
  data via risk-seeking policy gradients.
\newblock In {\em International Conference on Learning Representations}, 2021.

\bibitem{mundhenk2021seeding}
T.~Nathan Mundhenk, Mikel Landajuela, Ruben Glatt, Claudio~P. Santiago,
  Daniel~M. Faissol, and Brenden~K. Petersen.
\newblock Symbolic regression via neural-guided genetic programming population
  seeding.
\newblock In {\em 35th Conference on Neural Information Processing Systems
  (NeurIPS 2021)}, 2021.

\bibitem{https://doi.org/10.48550/arxiv.2205.11798}
Jiachen Li, Ye~Yuan, and Hong-Bin Shen.
\newblock Symbolic expression transformer: A computer vision approach for
  symbolic regression, 2022.

\bibitem{SciNet}
Raban Iten, Tony Metger, Henrik Wilming, L\'{\i}dia del Rio, and Renato Renner.
\newblock Discovering physical concepts with neural networks.
\newblock {\em Phys. Rev. Lett.}, 124:010508, Jan 2020.

\bibitem{pmlr-v139-biggio21a}
Luca Biggio, Tommaso Bendinelli, Alexander Neitz, Aurelien Lucchi, and
  Giambattista Parascandolo.
\newblock Neural symbolic regression that scales.
\newblock In Marina Meila and Tong Zhang, editors, {\em Proceedings of the 38th
  International Conference on Machine Learning}, volume 139 of {\em Proceedings
  of Machine Learning Research}, pages 936--945. PMLR, 18--24 Jul 2021.

\bibitem{NIPS2017_3f5ee243}
Ashish Vaswani, Noam Shazeer, Niki Parmar, Jakob Uszkoreit, Llion Jones,
  Aidan~N Gomez, \L~ukasz Kaiser, and Illia Polosukhin.
\newblock Attention is all you need.
\newblock In I.~Guyon, U.~V. Luxburg, S.~Bengio, H.~Wallach, R.~Fergus,
  S.~Vishwanathan, and R.~Garnett, editors, {\em Advances in Neural Information
  Processing Systems}, volume~30. Curran Associates, Inc., 2017.

\bibitem{vae-rnn}
Samuel~R. Bowman, Luke Vilnis, Oriol Vinyals, Andrew Dai, Rafal Jozefowicz, and
  Samy Bengio.
\newblock Generating sentences from a continuous space.
\newblock In {\em Proceedings of The 20th {SIGNLL} Conference on Computational
  Natural Language Learning}, pages 10--21, Berlin, Germany, August 2016.
  Association for Computational Linguistics.

\bibitem{popov_sergei_2022_7364439}
Popov Sergei, Lazarev Mikhail, Belavin Vladislav, Derkach Denis, and
  Ustyuzhanin Andrey.
\newblock {Symbolic expression generation via Variational Auto-Encoder},
  November 2022.

\bibitem{kingma2013auto}
Diederik~P Kingma and Max Welling.
\newblock Auto-encoding variational bayes.
\newblock {\em arXiv preprint arXiv:1312.6114}, 2013.

\bibitem{DL4SM}
Guillaume Lample and François Charton.
\newblock Deep learning for symbolic mathematics.
\newblock In {\em International Conference on Learning Representations}, 2020.

\bibitem{bfgs}
R.~Fletcher.
\newblock {\em Practical Methods of Optimization}.
\newblock John Wiley \& Sons, Ltd, 1987.

\end{thebibliography}

\end{document}